\pdfoutput=1

\documentclass[11pt]{article}

\usepackage[preprint]{acl}

\usepackage{times}
\usepackage{latexsym}

\usepackage[T1]{fontenc}

\usepackage[utf8]{inputenc}

\usepackage{microtype}

\usepackage{inconsolata}

\usepackage{graphicx}

\usepackage{array}

\usepackage{tabularx}
\usepackage{subfigure,graphicx}
\usepackage{booktabs}
\usepackage{multirow}
\usepackage{makecell}
\usepackage{xcolor}
\usepackage{eqparbox} 
\usepackage{transparent}
\usepackage{subcaption} 
\usepackage{hyperref}
\usepackage[most]{tcolorbox}
\newcommand{\ie}{\textit{i.e.}}
\newcommand{\eg}{\textit{e.g.}}

\newcommand{\ours}{{RealVideoQuest}}

\title{Respond Beyond Language: A Benchmark for Video Generation in Response to Realistic User Intents}

\author{Shuting Wang$^{1,2,\dagger}$, Yunqi Liu$^{1,\dagger}$, Zixin Yang$^{1}$, Ning Hu$^{3}$, Zhicheng Dou$^{2}$, Chenyan Xiong$^{1,*}$  \\
$^1$School of Computer Science, Carnegie Mellon University \\
$^2$Gaoling School of Artificial Intelligence, Renmin University of China \\
$^3$Serendipity One Inc. \\
\texttt{\{shutingw, yunqiliu, cx\}@andrew.cmu.edu}
}

\begin{document}
\maketitle
\def\thefootnote{*}\footnotetext{Corresponding author.}
\def\thefootnote{$\dagger$}\footnotetext{Equal contribution.} 
\def\thefootnote{\arabic{footnote}}

\begin{abstract}

Querying generative AI models, \eg, large language models (LLMs), has become a prevalent method for information acquisition. However, existing query-answer datasets primarily focus on textual responses, making it challenging to address complex user queries that require visual demonstrations or explanations for better understanding.
To bridge this gap, we construct a benchmark, \ours{}, designed to evaluate the abilities of text-to-video (T2V) models in answering real-world, visually grounded queries. It identifies 7.5K real user queries with video response intents from Chatbot-Arena and builds 4.5K high-quality query-video pairs through a multistage video retrieval and refinement process. We further develop a multi-angle evaluation system to assess the quality of generated video answers. 
Experiments indicate that current T2V models struggle with effectively addressing real user queries, pointing to key challenges and future research opportunities in multimodal AI.

\end{abstract}

\section{Introduction}
Generative AI models, particularly LLMs, have significantly transformed information acquisition ways by allowing users to issue natural language queries and receive generated answers directly. 
However, current query-answering tasks are limited to textual responses~\cite{nq,yang-etal-2018-hotpotqa,domainrag,omnieval}, 
overlooking scenarios where complex queries demand more than just textual answers. In many cases, visual demonstrations can significantly enhance user comprehension on responses and facilitate problem-solving. 
For example, domains such as skill learning often demand video responses to adequately satisfy user information needs. Unfortunately, existing text-to-video datasets~\cite{webvid,openvid} primarily consist of paired video-text descriptions, neglecting the task of answering real user queries with meaningful visual content.

To address this gap, we develop a novel query-to-video benchmark, \ours{}, designed to assess the capabilities of text-to-video generation models in answering real-world, complex user queries. Specifically, we curate 7.5K user queries that demand video-format responses, sourced from authentic user interactions on Chatbot-Arena~\footnote{\hyperlink{https://lmarena.ai/}{https://lmarena.ai/}}. For each query, we retrieve the top-1 long video from YouTube~\footnote{\hyperlink{https://www.youtube.com/}{https://www.youtube.com/}} and extract the most relevant clips to form video answers. We further apply a query rewriting process to better align the user intent with video answers, resulting in a refined and high-quality dataset of query-video answer pairs.

Our evaluation system defines four metrics, relevance, correctness, coherence, and completeness, to assess how well the generated videos address user queries. Combined with existing video quality evaluation methods~\cite{vbench,he2024videoscore}, we build a multi-angle evaluation system to comprehensively evaluate the T2V models on our challenging task. 

We evaluate several promising models, including T2V-Turbo-v2~\cite{t2v-turbo-v2}, CogVideoX-5B~\cite{yang2024cogvideox}, Hunyuan~\cite{kong2024hunyuanvideo}, SkyReels~\cite{SkyReelsV1}, and Wan2.1~\cite{wan2025}. Although these models can generate visually appealing videos from text prompts, our results indicate that they struggle to accurately and sufficiently answer user queries requiring visual demonstrations in practice. 
We attribute this problem to two key limitations: 
the lack of structured world knowledge and the difficulty in generating long and coherent video content. This problem also merits further exploration in future research on text-to-video generation.

\section{Construction of \ours{}}

In this section, we illustrate the construction pipeline of our benchmark, \ours{}. 
\subsection{Collection of Real User Queries}\label{sec:query_collection}
\paragraph{Collection of real user intents.} To ensure the authenticity of our dataset, we collect real queries with video generation intents from two realistic human-AI conversation datasets: LMSYS-Chat-1M~\footnote{\href{https://huggingface.co/datasets/lmsys/lmsys-chat-1m}{lmsys-chat-1m}} and Chatbot-Arena~\footnote{\href{https://huggingface.co/datasets/lmsys/chatbot_arena_conversations}{chatbot\_arena\_conversations}}, both released from the Chatbot Arena platform. We filter out non-English user queries and result in 800K real user queries, encompassing various types of real user intent during conversations with LLMs, allowing us to analyze the distribution of video-intent user queries.

\paragraph{Video intent recognition.} Subsequently, we build a video intent recognizer (VIR) based on GPT-3.5-Turbo~\cite{GPT-3.5}, which targets to identify user queries that desire video-format responses from all arena queries. 
Specifically, we define three-scale labels for query identification: ``2'' means that a video format answer is better than a textual answer, since it conveys more vivid and clear information than text; ``1'' indicates that a video format or a textual answer is suitable for answering the question; ``0'' means that the textual answer is better than a video format answer, which may be because the query distinctly requires a textual answer, such as ``writing a poem'', etc. Furthermore, we treat queries labeled as ``2'' and ``1'' as video-intent queries to expand the amount of our dataset. 
All prompts we used in our study are presented in Appendix~\ref{app:prompts}. 
Finally, we gather 7.5K real video-intent queries from all collected data.

\subsection{Categorization of User Intents}\label{sec:query_categorize}

Our preliminary user study categorizes video-intent queries into four main types:
(1) \textit{Skill demonstration}:
Users seek instructional videos for learning practical tasks, such as ``How to make a cupcake,'' where step-by-step visual guidance is crucial.
(2) \textit{Knowledge explanation}:
For complex concepts that are difficult to convey through text, users request visualizations, \eg, ``Show a visual breakdown of how the human circulatory system works.''
(3) \textit{Art creation}:
Beyond typical text-to-video mappings~\cite{webvid, nan2024openvid}, these queries involve creative problem-solving, like ``Make a funny commercial for a Honda Civic starring Allen Iverson,'' demanding original, design-driven video content.
(4) \textit{Human-machine interaction}:
Leveraging the capabilities of generative models, users engage in interactive tasks that require visually grounded interactive responses, \eg ``Let's play Gobang''.
We use GPT-3.5-Turbo to classify our identified video-intent queries. More details are provided in Appendix~\ref{app:query_type}.
\subsection{Building Video Answers for Queries}\label{sec:qa-pair}
Given the collected video-intent queries, we design a multi-stage pipeline to obtain their video answers, thereby supporting the subsequent evaluation. 
\paragraph{Retrieving relevant videos.}
First, we systematically retrieve top-1 high-resolution videos from YouTube using our video-intent queries. 

\paragraph{Video Splitting.} 
Since the retrieved YouTube videos are lengthy and information-overloaded, we further split these long videos into meta-clips with complete semantics. Following Panda-70M~\cite{chen2024panda70m}, the video splitting is implemented by PySceneDetect~\footnote{\hyperlink{https://www.scenedetect.com/}{https://www.scenedetect.com/}}. Each clip is then encoded into a multi-modal representation by the ImageBind model~\cite{girdhar2023imagebind}. 
To ensure temporal and semantic coherence, adjacent clips with cosine similarity exceeding the predefined threshold (0.3) are merged into cohesive segments. The lengths of the final video segments are around 15–60 seconds. 

\paragraph{Query-Video Alignment}

For each query, we only retain one of the video segments with the highest semantic similarity with the query to build the query-video answer pair with high relevance. To measure such similarity, we first generate textual descriptions for each video clip through Qwen2VL-7B~\cite{Qwen2-VL}, then compute cosine similarity scores between the query and clip descriptions using the BGE-large-en-v1.5~\cite{bge_embedding}. 
Furthermore, since the retrieved videos are refined into more detailed and specific video segments, we also rewrite original queries using GPT-4o~\cite{openai2024gpt4technicalreport}, conditioned on the video segment, to make the rewritten queries more specific and better aligned with video answers. 

We present the final statistical information of \ours{} in Table~\ref{tab:statis}.

\begin{table*}[h]
\caption{Overall performance of existing T2V models on our video answer evaluation. All results are normalized from [0, 4] to [0, 1] by dividing by 4. The best and the second-best results are highlighted in \textbf{bold} and \underline{underline}.}
\resizebox{0.99\linewidth}{!}{
\begin{tabular}{lcccccccccc}
\toprule
\multirow{2}{*}{Models} & \multicolumn{5}{c}{Non-reference-based Evaluation}            & \multicolumn{5}{c}{Reference-based Evaluation}                \\
\cmidrule(lr){2-6}\cmidrule(lr){7-11}
& Relevance & Correctness & Coherence & Completeness & AVG. & Relevance & Correctness & Coherence & Completeness & AVG. \\
\midrule
T2V-Turbo-V1      & 0.3311 & 0.2795 & 0.3979 & 0.2247 & 0.3083 & 0.2154 & 0.1804 & 0.2711 & 0.1476 & 0.2036    \\
T2V-Turbo-V2      & 0.3634 & 0.3228 & 0.4377 & 0.2505 & 0.3436 & 0.2242 & 0.2022 & 0.3019 & 0.1586 & 0.2217    \\
CogVideoX-5B      & 0.2315 & 0.2105 & 0.3006 & 0.1587 & 0.2253 & 0.1295 & 0.0974 & 0.1899 & 0.0747 & 0.1229    \\
Hunyuan           & \underline{0.3780} & \underline{0.3562} & \underline{0.4587} & \underline{0.2806} & \underline{0.3684} & \underline{0.2721} & \underline{0.2573} & \underline{0.3612} & \underline{0.1989} & \underline{0.2724}    \\
SkyReels          & 0.3404 & 0.3076 & 0.4064 & 0.2408 & 0.3238 & 0.2455 & 0.2285 & 0.3184 & 0.1729 & 0.2413   \\
Wan2.1            & \textbf{0.3909} & \textbf{0.3569} & \textbf{0.4800} & \textbf{0.2876} & \textbf{0.3788} & \textbf{0.2740} & \textbf{0.2740} & \textbf{0.3923} & \textbf{0.2134} & \textbf{0.2885}    \\
\bottomrule
\end{tabular}
}
\label{tab:overall_our}
\end{table*}
\begin{table*}[!h]
\caption{Overall performance on VideoScore and practicable evaluation dimensions of VBench. }
\resizebox{0.99\linewidth}{!}{
\begin{tabular}{lccccccccccccc}
\toprule
\multirow{2}{*}{Models} & \multicolumn{6}{c}{VideoScore} & \multicolumn{7}{c}{Vbench}  \\
\cmidrule(lr){2-7}\cmidrule(lr){8-14}
& \makecell[c]{Visual\\Quality} & \makecell[c]{Temporal\\Consistency} & \makecell[c]{Dynamic\\Degree} & \makecell[c]{Text-to-video\\Alignment} & \makecell[c]{Factual\\Consistency} & AVG. & \makecell[c]{Image\\Quality} & \makecell[c]{Aesthetic\\Quality} & \makecell[c]{Dynamic\\Degree} & \makecell[c]{Motion\\Smoothness} & \makecell[c]{Background\\Consistency} & \makecell[c]{Subject\\Consistency} & AVG. \\ 
\midrule
T2V-Turbo-V1     & 2.6103 & 2.3555 & 2.7490 & \underline{2.5552} & 2.2629 & 2.5066 & \textbf{73.46} & \textbf{58.50} & 29.60 & 97.04 & 96.88 & \textbf{97.91} & 75.57  \\
T2V-Turbo-V2     & 2.4458 & 2.3540 & 2.7805 & 2.4192 & 2.3601 & 2.4719 & \underline{71.71} & 53.22 & \textbf{73.50} & 98.11 & 95.08 & 95.48 & \textbf{81.18}  \\
CogVideoX-5B     & 2.7018 & 2.3638 & \underline{2.8102} & 2.1221 & 2.4778 & 2.4951 & 52.93 & 41.76 & 22.70 & \underline{99.33} & 94.16 & 90.51 & 66.90  \\
Hunyuan          & \underline{2.9489} & \underline{2.7427} & 2.7618 & 2.4347 & \underline{2.8245} & \underline{2.7425} & 71.49 & 54.05 & 28.16 & \textbf{99.57} & \textbf{97.38} & \underline{97.61} & 74.71  \\
SkyReels         & \textbf{3.4044} & \textbf{3.1988} & \textbf{3.3616} & \textbf{2.9933} & \textbf{3.3168} & \textbf{3.2550} & 61.76 & 46.02 & 35.80 & 99.17 & \underline{97.31} & 96.90 & 72.83  \\
Wan2.1           & 2.7629 & 2.4866 & 2.7214 & 2.4698 & 2.4758 & 2.5833 & 71.04 & \underline{56.82} & \underline{60.30} & 99.01 & 96.89 & 96.25 & \underline{80.05}  \\
\bottomrule
\end{tabular}
}
\label{tab:overall_raw}
\end{table*}
\section{Multi-angle Evaluation System}\label{sec:evaluation}
We find that existing text-to-video evaluation suites~\cite{vbench, internvid} are insufficient for our query-answer (QA) format task, as they primarily focus on visual quality and the consistent matching between input descriptions and generated videos.
To effectively assess the quality of generated video answers, we propose four QA-quality metrics:
(1) \textit{Relevance}: Assesses topic alignment between the query and the video;
(2) \textit{Correctness}: Measures how accurately the video addresses the query;
(3) \textit{Coherence}: Evaluates the logical consistency of the video's progression;
(4) \textit{Completeness}: Determines whether the video fully resolves the query task.
Each metric is rated on a scale from 0 (lowest) to 4 (highest).

Given the strong instruction-following and video understanding abilities of multimodal large language models (MLLMs), we adopt the LLM-as-a-Judge method using GPT-4o-mini, with a two-branch evaluation strategy:
(1) \textit{Non-reference-based}: Directly inputs instructions, queries, and generated videos into the MLLM to evaluate QA quality, utilizing the inherent capabilities of MLLMs.
(2) \textit{Reference-based}: Further includes golden video answers to guide the evaluation.
\begin{table}[t]
\caption{Query statisitc of \ours{}.}
\resizebox{0.99\linewidth}{!}{
\begin{tabular}{l@{}rrrr}
\toprule
Dataset                   & All Queries  & \makecell[c]{All QA pairs} & Training & Test \\
\midrule
Skill demonstration       & 1,381 & 1,024             & 827      & 197  \\
Knowledge explanation     & 2,833 & 1,820             & 1,433     & 387  \\
Human-machine interaction & 708  & 263              & 195      & 68   \\
Art creation              & 508  & 258              & 197      & 61   \\
Else                      & 2,101 & 1,246             & 961      & 285  \\
Sum                       & 7,531 & 4,611             & 3,613     & 998  \\
\bottomrule
\end{tabular}
}
\vspace{-8pt}
\label{tab:statis}
\end{table}

We also use VBench~\cite{vbench} and VideoScore~\cite{he2024videoscore} to measure video visual quality, building our multi-angle evaluation system. For VBench, we choose universally applicable dimensions: image quality, aesthetic quality, dynamic degree, motion smoothness, background consistency, and subject consistency, excluding those that need prompt-specific meta-information.\footnote{\hyperlink{https://github.com/Vchitect/VBench}{https://github.com/Vchitect/VBench}} VideoScore assesses videos from visual quality, temporal consistency, dynamic degree, text-to-video alignment, and factual consistency and predicts scores ranging in $[1,4]$ for each dimension.\footnote{\hyperlink{https://huggingface.co/TIGER-Lab/VideoScore-v1.1}{https://huggingface.co/TIGER-Lab/VideoScore-v1.1}} The prompts for LLM-as-a-Judge are presented in Appendix~\ref{app:prompts}. We will publish the data construction and evaluation codes upon acceptance of our study.
\begin{table}[t]
    \centering
    \small
    \caption{Accuracy of our video intent recognizer.}
    \begin{tabular}{>{\centering\arraybackslash}p{1.3cm}>{\centering\arraybackslash}p{1.3cm}>{\centering\arraybackslash}p{1.3cm}>{\centering\arraybackslash}p{1.3cm}}
    \toprule
    Accuracy & Precision & Recall & Cost(\$) \\
    \midrule
    0.87 & 0.2778 & 1.0000 & 0.0015 \\
    \bottomrule
    \end{tabular}
    \vspace{-5pt}
    \label{tab:vir_acc}
\end{table}
\section{Experiment}
\begin{figure*}[t]
    \centering
    \includegraphics[width=0.85\textwidth]{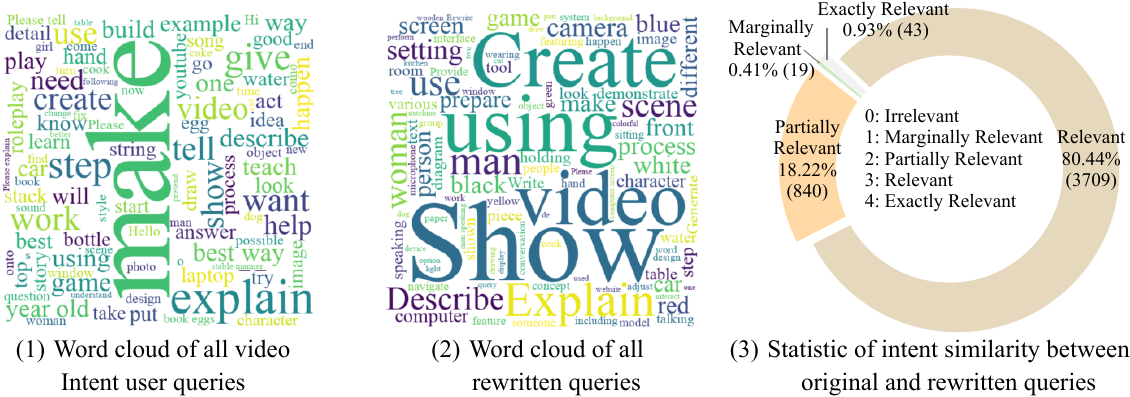}
    \caption{The word clouds of original and rewritten queries and the visualization of their statistical similarity.} %
    \label{fig:query_rewrite}
\end{figure*}
We provide a holistic evaluation of existing T2V models on our benchmark and further analyses. 
\subsection{Performance of Current T2V Models}
Given our test queries, we infer various open-source T2V models performing well on VBench, and assess both their QA and video generation capabilities. The baselines include T2V-Turbo-V1 and V2~\cite{t2v-turbo-v1,t2v-turbo-v2}, CogVideoX-5B~\cite{yang2024cogvideox}, HunyuanVideo~\cite{kong2024hunyuanvideo}, SkyReels-V1 (SkyReels for short)~\cite{SkyReelsV1}, a fine-tuned and faster version of HunyuanVideo, and Wan2.1~\cite{wan2.1}. 
Since video generation is time-consuming, we yield one video per query, which may introduce some randomness. The results are shown in Tables~\ref{tab:overall_our} and~\ref{tab:overall_raw}. 

Overall, while current T2V models generate visually impressive outputs with some metric values, \eg motion smoothness and subject consistency, nearing saturation, they struggle to effectively address user queries, as reflected by low scores on our QA-quality metrics. Among these, completeness proves especially challenging, likely due to the short duration (typically a few seconds) of generated videos. 
These findings validate our motivation: 
despite strong performance on traditional benchmarks, existing T2V models fall short in handling realistic, visually grounded user tasks, which highlights a crucial direction for future research.

\subsection{Further Analyses}

\paragraph{Quality of video intent recognizer.}
To test the quality of VIR, we employ two annotators to label 100 arena queries whether they exhibit video intents, \ie, label $\geq1$. The Cohen’s Kappa of the annotation is 0.4973, indicating moderate agreement and reliable annotations. 
By treating a query as video-intent if at least one annotator marked it as such, we aggregate their labels to form the final test set. The evaluation result of our VIR are shown in Table~\ref{tab:vir_acc}. 
It indicates that VIR achieves strong recall and overall accuracy, though its precision is relatively limited. However, since the downstream retrieval can inherently filter out unsuitable queries, we prioritize recall over precision in this stage.
\paragraph{Consistency of rewritten queries}
We also test the consistency between rewritten queries and original ones to prove that our query rewriter could maintain original user intents while aligning with video answers. We first present the word clouds of original and rewritten query sets in Figures~\ref{fig:query_rewrite}. Noticeably, some words, \eg, ``make'' (``create''), ``explain'',  and``step''(``process''), are high-frequency on two query sets, implying that both query sets mainly contain users' practical requests for AI models. We then use GPT-3.5-Turbo to judge the intent similarity between original and rewritten queries using a 5-point Likert scale ranging from 0 to 4, where 0 indicates no similarity and 4 represents exact relevance. 
Considering the diverse presentations~\cite{PRADA} of queries, we first extract key topics from queries and then identify topic similarity between two queries, ensuring a robust and reliable identification of query similarity. 
The statistical results are shown in Figure~\ref{fig:query_rewrite} (c). Evidently, almost all rewritten queries exhibit high relevance (3) to original queries, proving the reliability of our rewriting module. 

Due to the limited space, we provide some case studies and analyses in Appendix~\ref{app:case}. 
\section{Conclusion}
In this study, we advanced the query-answering task from textual to video-based responses by creating a new benchmark, \ours{}. It gathers real user queries with video-answer intent from ChatbotArena and develops a multistage data process to retrieve and create high-quality query-video pairs. We also built a multi-angle evaluation framework by combining our QA-quality metrics with existing video quality metrics. 
Experimental results validate the motivation and importance of \ours{}, revealing that current text-to-video generation models struggle to adequately address real user intents, highlighting promising directions for future text-to-video research.

\newpage
\section*{Limitation}
We introduces \ours{}, a novel benchmark for evaluating T2V models on responding to real-world, visually grounded user queries. It extracts 7.5K real queries with video-answer intent from the ChatbotArena dataset and constructs 4.5K high-quality query-video pairs via a multistage retrieval and refinement pipeline. Furthermore, we propose a multi-angle evaluation framework that combines fine-grained QA-quality metrics with established visual quality assessments, enabling comprehensive analysis of T2V model capabilities.

While \ours{} focuses on real-world queries with visual response intent, it is constrained by the quality and diversity of retrieved YouTube videos, which may not fully represent the ideal responses users expect from generative T2V systems. Additionally, due to the high computational cost of video generation, we evaluate only a single generated video per query, which may not capture the full variability or potential of each model. These factors limit the completeness and generalizability of our current evaluation.

\section*{Ethical Statements}
In this paper, we develop a QA-oriented task for the text-to-video generation. To build reliable query-video answer pairs, we retrieve video answers from YouTube and clip video segments to form video answers. To ensure the legitimacy of our research dataset, we will only publicize YouTube URLs with their start and end timestamps to provide indirect video information.



\newpage
\bibliography{custom}

\appendix
\appendix
\begin{figure*}
    \centering
    \includegraphics[width=1\linewidth]{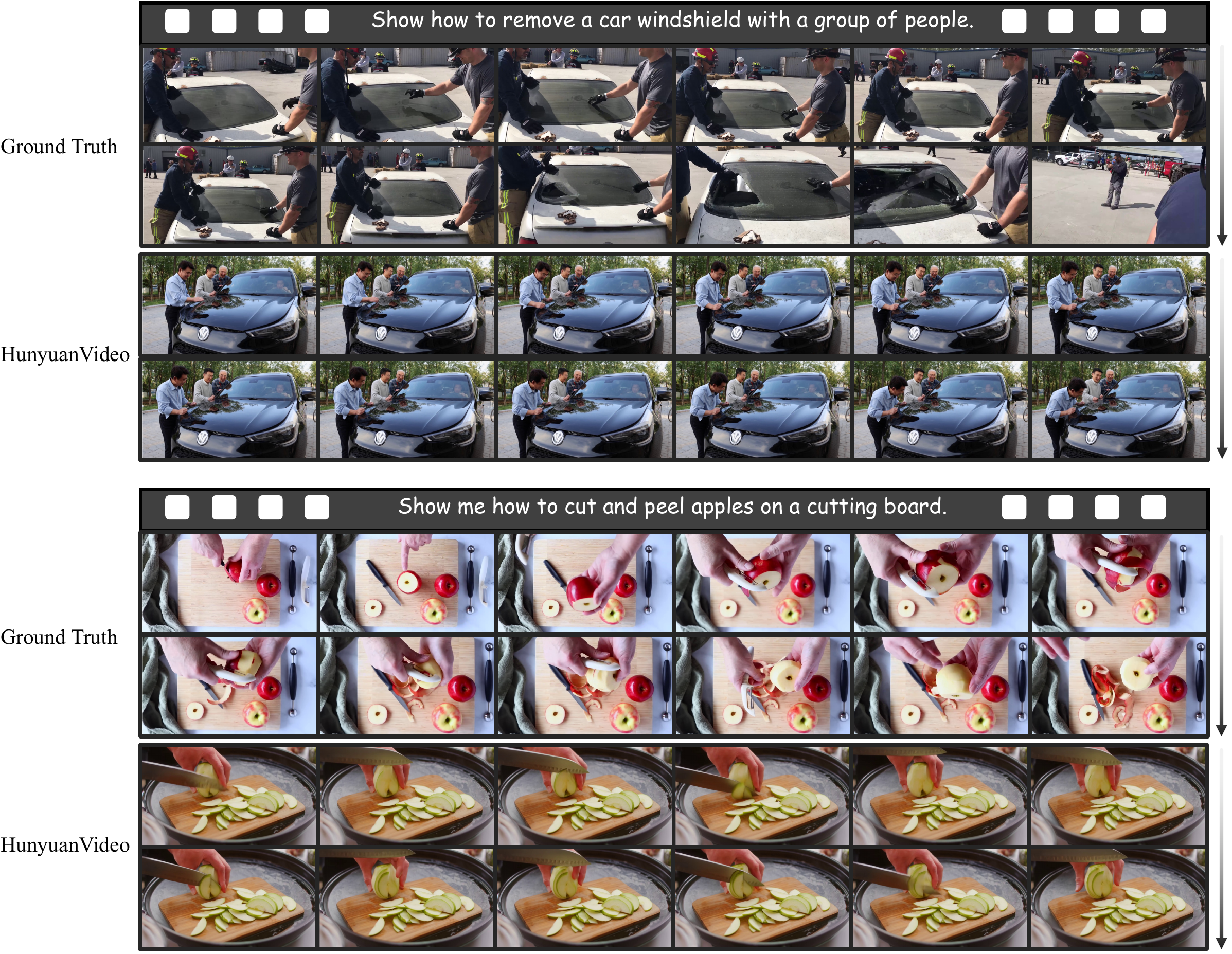}
    \caption{Comparison between generated videos and ground truths.}
    \label{fig:case}
\end{figure*}
\section{Detailed Descriptions of Query Types and Examples}\label{app:query_type}
In this section, we further provide the detailed descriptions of our self-devised query types and the corresponding query examples in Table~\ref{tab:query_type}.
\begin{table*}[]
    \centering
    \caption{Definition and examples of four types of video intent queries.}
    \resizebox{0.99\linewidth}{!}{\begin{tabular}{p{4cm}p{7.5cm}p{4.5cm}}
    \toprule
    Type & Definition & Examples \\
    \midrule
    Skill Demonstration & Question that asks for demonstrating skills, such as cooking, paper folding, car repairing, and so on. Text alone might be limited in its instructional ability where demonstrations are desired. & ``Explain how to tie a knot''\newline ``How to make a pizza''\\
    \midrule
    Knowledge Explanation.  & Questions related to knowledge-intensive concepts or entities that are better explained with graphics or animation. These are typically complex concepts or entities where visual aids clarify relationships, processes, or dynamic phenomena better than static text alone. & ``Describe how a volcano works to a five year-old. Answer:''\newline ``How do helicopters fly?'' \\
    \midrule
    Art Creation and Design & Question that explicitly asks for creating or design images, video, animation, and so on. & ``Can you create a human, male character based on the Bandersnatch?''\\
    \midrule
    Human-machine interaction & Questions that request AI models to interacte with users. & ``Let's play  a game of tic-tac-toe. You go first'' \\
    \bottomrule 
    \end{tabular}}
    \label{tab:query_type}
\end{table*}

\section{Task Prompts Used in Our Study}\label{app:prompts}
In this section, we provide the detailed prompts for all tasks in our study, including video intent classification (Table~\ref{tab:prompt_video_intent}), video caption generation, query rewriting (Table~\ref{tab:prompt_video_caption_qr}), and identification of query intent similarity (Table~\ref{tab:query_similarity}). We also provided the template and detailed descriptions for each metric used for our LLM-as-a-Judge evaluation method in Figure~\ref{box:eval_template} and Table~\ref{tab:metric_desc}. 

\begin{table*}[]
    \centering
    \caption{Prompts for Video Intent Classification.}
    \label{tab:prompt_video_intent}
    \resizebox{0.99\linewidth}{!}{%
    \begin{tabular}{p{4cm}p{15cm}}
    \toprule
    Task & Prompt\\
    \midrule
    Video Intent Recognition (first-round filtering) &

    You will be provided with a user query to a generative model.\newline
    Please judge whether the query can be answered via a video.\newline
    Return 1 if it can be answered via video. Return 0 otherwise.\newline
    Your response should only be a number 0 or 1.\\
    
    \midrule
    Video Intent Recognition (second-round filtering) 
    &
          \# Task Description\newline
          You will be provided with a user query directed at a generative model.
          Your task is to determine whether the query would be better answered via a video rather than text.\newline
          \# Guidelines\newline
          1. If the query requires visual demonstrations, dynamic processes, or relies on visual or auditory context, it should be answered via video.\newline
          2. If the query can be fully and clearly answered using concise text, numbers, or static information, it should not be answered via video.\newline
          \# Instructions\newline
          Return 1 if the query would be better answered via a video.\newline
          Return 0 otherwise.\newline
          Your response should only be a number 0 or 1.\\

    \midrule
    Video Intent Recognition (third-round filtering) &
    \# Task Description\newline
    You will be provided with a user query directed at a generative model.\newline
    Your task is to determine whether the query is an instruction.\newline
    \# Guidelines\newline
    A query is an instruction if the user asks for an answer, guidance, or asks a question.\newline
    Think carefully.\newline
    Return 1 if the query is an instruction. Return 0 otherwise.\newline
    Your response should only be a number 0 or 1. \\
    \bottomrule
    \end{tabular}}
\end{table*}
\begin{table*}[]
    \centering
    \caption{Prompts for Video Caption Generation and Query Rewrite}
    \label{tab:prompt_video_caption_qr}
    \resizebox{0.99\linewidth}{!}{%
    \begin{tabular}{p{4cm}p{15cm}}
    \toprule
    Task & Prompt\\
    \midrule
    Clip Description Generation
    &
    \# Task Description \newline
    You will be provided with a short video or its keyframes.\newline
    Your task is to generate a concise and descriptive caption that summarizes the overall content\newline of the video, not just the beginning scene.\newline
    
    \# Guidelines\newline
    - Consider the entire video when generating the caption.\newline
    - If the video contains text or spoken words, explicitly mention that the video contains words\newline and briefly describe their content.\newline
    
   \# Instructions\newline
    Write a clear and informative single-sentence caption that accurately reflects the main \newline
    content and context of the video.\\
    
    \midrule
    Query Rewrite
    &
    I want to create an instruction tuning dataset for text to video generative model. I use a \newline
    query to fetch related video from Youtube, and I want to rewrite that query based on the \newline
    content of the video to make the query more aligned with the video. Rewrite the query for me. \newline
    Your response should only be one sentence, and similar to the original query, it should contain
    \newline
    what a person asks the model to do. The original query is: \{original query\}, and the video description is \{video description\}"\\
    \midrule
    Query Type Classification & \#\# Background \newline
    You are a classifier that determines which category a query belongs to.\newline
    Here are the categories and examples:\newline
    1. Art creation and designing\newline
    Example: "generate a unique design of LED light for house"\newline

    2. Skill demonstration\newline
    Example: "How do i clean my water bottle if i can't reach down into it", "How to bake a cake?"\newline

    3. Knowledge explanation\newline
    Example: "how sun makes energy?", "hello, give me a short visual description of The Fool tarot card"\newline

    4. Human-machine interaction and role play\newline
    Example: "Pretend you are Spiderman and wish me for my birthday"\newline

    \#\# Output format\newline
    Return only a number from 0 to 4, where 1-4 correspond to the given categories, and 0 means the query does not fit into any category.
    Do not return anything other than a number.\newline
    
    Query: \$\{query\}\newline Return only a number from 0 to 4. \\
    \bottomrule
    \end{tabular}}
\end{table*}
\begin{table*}[]
    \centering
    \caption{Prompts for Similar Query Intent Recognition.}
    \label{tab:query_similarity}
    \resizebox{0.99\linewidth}{!}{%
    \begin{tabular}{p{4cm}p{15cm}}
    \toprule
    Task & Prompt \\
    \midrule
    Identify Query Topics & You are a researcher analyzing user queries to summarize their essential demands.\newline
    Your task is to:\newline
    - If the query contains multiple requests or needs, break them into key points.\newline
    - Only return as many needs as necessary.\newline
    - Return at most two needs.\newline
    
    Examples:\newline
      Query: How do I improve my website's SEO ranking and optimize loading speed?\newline
      Summary: [\newline
          \quad"SEO ranking improvement",\newline
          \quad"Website loading speed optimization"\newline
        ]\newline
      Query: Recommend a 30 minute workout for weight loss that includes jump roping and interval training for a 30 year old man that exercises often 3-4 days a week who has access to a full gym.\newline
      Summary: [\newline
            \quad"Workout plan recommendation for weight loss"\newline
        ]\newline
    
    Provide the summary as a **Python list**.\newline
    
    Query: "\$\{query\}"\newline
    Insights: \\
    \midrule
    Calculate Topic Similarity
    & \#\# Background \newline
    You are given two queries and the two corresponding lists of topics they contain. Analyze the overall semantic similarity between the new topics and the old topics based on the content and the meaning of the topics.\newline
    Note that the topics and the queries don't have to have identical or similar wording to be considered similar, they can be considered identical as long as the meaning is the same.\newline
    Your task is to output only an integer from 0 to 4, representing the similarity level:\newline
    0: completely unrelated\newline
    1: weakly related\newline
    2: somewhat related\newline
    3: strongly related\newline
    4: almost identical topics\newline
    
    Return only the integer. No explanation. \\
    \bottomrule
    \end{tabular}}
\end{table*}

\begin{figure*}[!h]
    \begin{tcolorbox}[title={The prompt template for our LLM-as-a-Judge evaluation.}] 
\#\# Task definition\newline
You are an expert query-video answer evaluator, and Your task is to evaluate whether the generated response questions can well answer the user's queries and solve the user's needs. I will provide you with the user query, the generated response video from a text-to-video generation model. I will also provide you with a ground truth video, which is one of the most correct video answers for the input query (retrieved from the Internet).\newline\newline
Please note that the ground truth video is only an assessment reference. It provides the correct answer to the current query, but sometimes the correct answer is not unique. Therefore, when you evaluate the response video, you can refer to the key and general knowledge provided in the ground truth video. At the same time, please also evaluate the response video based on your own world knowledge.\newline\newline
Specifically, you should evaluate the response video from the following dimensions:
\$\{metric\_name\}\newline
\#\# Input information\newline
- Query: it is a user query issued to LLMs to expect a video answer. It is a sentence.\newline
- Ground truth video: the most correct video answer for the input query. It is provided as a set of images that capture key frames in the video.\newline
- Response video: the video generated by a text-to-video generation model from input query. You need to evaluate the quality of response video by referring to the ground truth video from the above three evaluation dimensions. This video is also provided as a set of images that capture key frames in the video.\newline\newline
\#\# Output requirements:\newline
Your returned output should be in the JSON format, which conforms to the following detailed format:
\$\{common\_output\}\newline
    \end{tcolorbox}
\caption{The prompt template for our LLM-as-a-Judge evaluation.}
\label{box:eval_template}
\end{figure*}

\begin{table*}[]
    \centering
    \caption{Metric description for LLM-as-a-Judge evaluation.}
    \label{tab:prompt_video_intent}
    \resizebox{0.99\linewidth}{!}{%
    \begin{tabular}{p{2cm}p{15cm}p{3cm}}
    \toprule
    Metric & Description & Output Requirment \\
    \midrule
    Relevance & Relevance: It measures whether the contained information of the response video is relevant to the input query. It is a four-scale rating with the introduction as below: \newline
 - 0 means the response video is totally unrelated to the input query.\newline
 - 1 means the response video contains slight relevance to the input query, but loses critical relevant information.\newline
 - 2 means the response video contains information that is fairly relevant to the input query, but contains a small amount of irrelevant information that is not fatal.\newline
 - 3 means the contained information in the response video is totally relevant to the query. & A int value that should be 0, 1, 2, or 3. It represents your rating result for the relevance of the response video. \\
 \midrule
 Correctness & Correctness: This metric measures the correctness of the response video, which is decided by assessing whether the response video correctly contains the key information for answering the query. It is a four-scale rating with the definition as below:\newline
 - 0 means the contained information in the response video is totally incorrect for answering the query.\newline
 - 1 means the response video partially contains some correct information for answering the query while violating key information. \newline
 - 2 means the response video partially contains the correct information that is critical for answering the query, while also violating a little nonfatal information. \newline
 - 3 means the information conveyed by the response video is totally correct and is critical for answering the query. & A int value that should be 0, 1, 2, or 3. It represents your rating result for the correctness of the response video. \\
 \midrule
 Coherence & This metric measures whether the development process or steps of the response video content are logical and consistent and whether the consistency is reasonable. It is a four-scale rating:  \newline
 - 0 means the content of the response video is totally non-coherent and illogical.  \newline
 - 1 means the content has a certain coherence and logics but has fatal logical errors.  \newline
 - 2 means the most logic of the response video is coherent, but have some nonfatal illogical problems. \newline
 - 3 means the development process or steps of the response video are totally logical, consistent, and coherent. & A int value that should be 0, 1, 2, or 3. It represents your rating result for the correctness of the response video. \\
 \midrule
 Completeness & Completeness: It evaluates the completeness of the response video and is a four-scale rating:\newline
 - 0 means the response video contains no useful information for answering the query.\newline
 - 1 means the response video answers a few aspects of the query, yet neglects some important aspects.\newline
 - 2 means the response video answers most aspects of the query, yet neglects a few nonfatal aspects.\newline
 - 3 means the response video completely and correctly answers the query. & A int value that should be 0, 1, 2, or 3. It represents your rating result for the completeness of the response video. \\
    \bottomrule
    \end{tabular}}
    \label{tab:metric_desc}
\end{table*}

\section{Case Study}\label{app:case}
We present some comparison cases between generated videos (HunyuanVideo as the representative) and ground truths to validate our motivation. To visualize the video content, we uniformly sample 12 frames for each video, and illustrate the cases in Figure~\ref{fig:case}. For the first case, the query intent is to demonstrate the way to remove a car windshield. However, even though the AI-generated video is specious and exhibits high visual quality, it conveys no useful information to satisfy the query need. While the ground truth video actually demonstrates the skill to remove (break) the windshield from the car. Similarly, for the second case, where the query requests to show the way to cut and peel apples, the ground truth video shows the whole process, while the generated video contains some content contrary to the facts (for the 8th frame, some apple slices appeared out of nowhere). These cases further indicate that the current T2V models lack critical world knowledge, therefore blocking their abilities to informatively respond to realistic user queries.



\end{document}